# Improving Layout Representation Learning Across Inconsistently Annotated Datasets via Agentic Harmonization

Renyu Li[*†]    Vladimir Kirilenko[*]    Yao You    Crag Wolfe

*Unstructured Technologies*

leah@unstructured.io

[*]Equal contribution. [†]Corresponding author.

### Abstract

Fine-tuning object detection (OD) models on combined datasets assumes annotation compatibility, yet datasets often encode conflicting spatial definitions for semantically equivalent categories. We propose an agentic label harmonization workflow that uses a vision-language model to reconcile both category semantics and bounding box granularity across heterogeneous sources before training. We evaluate on document layout detection as a challenging case study, where annotation standards vary widely across corpora. Without harmonization, naïve mixed-dataset fine-tuning degrades a pretrained RT-DETRv2 detector: on SCORE-Bench, which measures how accurately the full document conversion pipeline reproduces ground-truth structure, table TEDS drops from 0.800 to 0.750. Applied to two corpora whose 16 and 10 category taxonomies share only 8 direct correspondences, harmonization yields consistent gains across content fidelity, table structure, and spatial consistency: detection F-score improves from 0.860 to 0.883, table TEDS improves to 0.814, and mean bounding box overlap drops from 0.043 to 0.016. Representation analysis further shows that harmonized training produces more compact and separable post-decoder embeddings, confirming that annotation inconsistency distorts the learned feature space and that resolving it before training restores representation structure.

## 1 Introduction

Object detection models serve as foundational components in many visual understanding systems. Pretrained detectors are frequently adapted to new domains by fine-tuning on additional datasets. When datasets share compatible annotation conventions, this approach can improve model performance and extend coverage to new data distributions.

In many practical settings, however, datasets that appear compatible at the label level still encode different assumptions about how objects should be spatially annotated. For example, one dataset may annotate a document paragraph as a coarse block including surrounding whitespace, while another may use fine-grained boxes aligned with OCR extraction boundaries. Similar discrepancies arise for titles, captions, forms, tables, and list structures (Figure 1 and Figure 2). Prior work on multi-source and cross-dataset learning suggests that models frequently absorb such systematic inconsistencies as dataset-specific signals rather than learning invariances that transfer across datasets [1–3]. As a result, annotation style itself becomes a hidden source of distribution shift.

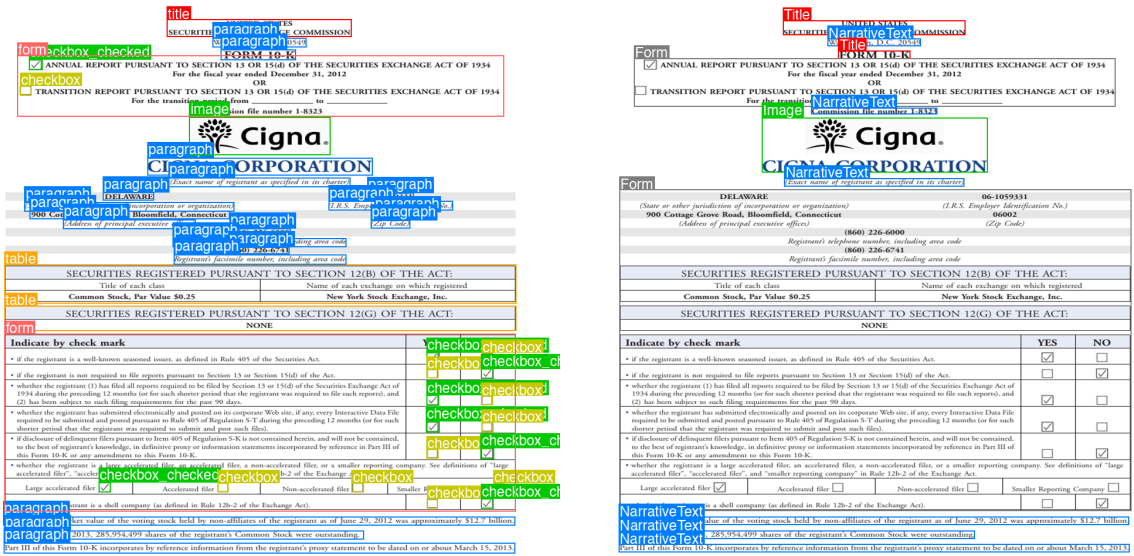

Figure 1: Annotation discrepancies - Tables and Forms

Figure 2: Annotation discrepancies - Paragraphs and List-Items

When heterogeneous datasets are combined through conventional fine-tuning, the detector receives conflicting supervision over semantically related categories. Even when label names appear compatible, the spatial priors encoded in the annotations may differ substantially. This makes it difficult for the model to learn a stable definition of object extent, leading to degraded recall, unstable localization, and weaker adaptation overall. Additional challenges arise when the new dataset differs significantly in size or distribution from the pretrained model's original training data, which can amplify data imbalance and catastrophic forgetting [4, 5].

This problem is especially acute in document layout detection, where annotation standards vary widely across datasets [6–8]. Document datasets often differ in both taxonomy and granularity because they are designed for different downstream goals, such as OCR, table extraction, or semantic segmentation. As a result, naïvely fine-tuning across such datasets can distort the detector's internal layout representation, even when the categories overlap.

In this work, we study cross-dataset adaptation for document layout detection under heterogeneous annotation semantics. We introduce an agentic label harmonization workflow that reconciles annotation differences before training. The agent leverages both visual and linguistic signals in two ways: 1) It analyzes each page image, grounding its spatial decisions in the underlying layout structure. 2) It reasons in natural language about annotation semantics: understanding what each label represents, how regions should be grouped, and where granularity conventions differ, in order to align both label meaning and bounding box granularity to a unified target standard, resulting in a more consistent supervision signal across datasets.

Using harmonized datasets, detectors can be fine-tuned without inheriting conflicting spatial priors. We evaluate this approach using a transformer-based document layout detector and show that harmonized training improves detection quality relative to both a pretrained baseline and naïve mixed-dataset fine-tuning. Beyond output-level metrics, we further show that harmonization changes the geometry of the learned feature space: ambiguous classes become more separable, clusters become more compact, and post-decoder representations exhibit clearer organization.

These results support a simple claim: cross-dataset document detection can fail not only because of appearance shift or label mismatch, but because annotation inconsistency corrupts the learned layout representation itself. By resolving these inconsistencies before training, harmonization restores representation structure and leads to better downstream detection.

This research makes the following contributions:

- We identify annotation inconsistency as a representation-level failure mode in cross-dataset document layout detection, showing that naïve mixed-dataset fine-tuning can distort the learned layout feature space.
- We introduce an agentic label harmonization workflow that aligns both label semantics and bounding box granularity across heterogeneous document datasets to produce a unified supervision signal prior to training.
- We demonstrate that harmonized training not only improves end-to-end document transformation quality, it also produces more structured and globally separable layout representations for ambiguous layout categories, supported by embedding visualizations and clustering-based analyses of the detector feature space.

## 2 Related Work

### 2.1 Object Detection Models

Object detection has evolved rapidly from two-stage region proposal pipelines [9] to efficient one-stage detectors [10–12]. More recently, transformer-based architectures have emerged that formulate detection as a set prediction problem [13–15]. While early transformers struggled with high computational costs, models such as RT-DETR [16] and RT-DETRv2 [17] have optimized this architecture for real-time applications.

In the document domain, these detection models have been supported by large-scale benchmarks such as PubLayNet [6], DocBank [7], and DocLayNet [8]. These works show that modern detectors can perform strongly when trained on large datasets with consistent annotations. More recent document layout models such as VGT [18], DocLayout-YOLO [19], and the Docling Heron models [20] further demonstrate the strength of detector-based layout analysis at scale. However, as models are trained on increasingly heterogeneous document corpora, the assumption of consistent supervision becomes increasingly fragile.

### 2.2 Cross-Dataset Object Detection

Adapting object detectors across datasets is commonly framed as a transfer learning or domain adaptation problem [21, 22]. A large body of work aligns feature distributions between source and target domains through adversarial training or related strategies [23–26]. These methods are effective when datasets share annotation conventions but differ in appearance or image statistics.

More recent work addresses the setting where multiple datasets are used jointly during training. Universal detection frameworks [27] aim to learn shared representations across heterogeneous sources, while large-vocabulary detection methods [28] expand the category set using weak supervision. At the same time, multi-source training introduces challenges such as data imbalance, inconsistent category coverage, and catastrophic forgetting [4, 5].

A key observation emerging from this literature is that datasets can provide conflicting supervision signals. While recent work on cross-dataset detection and annotation mismatch [2, 3, 29] has begun to highlight the negative impact of inconsistent supervision, most approaches treat these inconsistencies as something to be absorbed during optimization rather than explicitly resolving them before training. Notably, LAT [29] is the closest contemporary effort that jointly addresses class-level and bounding-box inconsistencies, but it does so during training via pseudo-labels and a privileged proposal generator, rather than rewriting the source annotations before training. This limitation becomes particularly severe when differences extend beyond label names to the spatial definition of objects, such as varying bounding box boundaries for identical semantic classes.

## 2.3 Dataset Harmonization and Label Alignment

Prior work has explored several forms of harmonization for heterogeneous supervision. In segmentation, this often involves manually or automatically merging label ontologies into a shared canonical taxonomy across datasets [1]. Recent grounded vision-language models [30, 31] further suggest a practical mechanism for language- and image-conditioned category alignment, while work on heterogeneous supervision transfer, such as LSDA [32] and open-vocabulary detection [33] address broader forms of supervision alignment.

Other work focuses on mitigating mismatches during training rather than correcting them beforehand. In the document domain, related efforts include source-free domain adaptation [34] and hierarchical document structure modeling through tree-based construction [35]. Large multi-source document layout models such as Heron [20] further motivates the need for methods that can reconcile heterogeneous annotations before detector adaptation.

Despite these advances, existing unification strategies operate primarily at the semantic level or address heterogeneous supervision in more indirect ways. They do not directly reconcile how semantically related objects are spatially instantiated across datasets, for example, whether a paragraph should be represented as a coarse text region or a set of finer OCR-aligned blocks. As a result, detectors trained on mixed document datasets may still inherit inconsistent spatial supervision. Our work differs in that it performs harmonization *before* training and explicitly aligns both category semantics and bounding box granularity.

## 2.4 Representation Quality in Document Understanding

Document understanding systems rely heavily on the quality of learned layout representations. Initial pre-training approaches such as LayoutLM [36], DocFormer [37], DiT [38], Donut [39], and LayoutLMv3 [40] established that structured visual-textual representations can transfer across diverse downstream tasks. Building on the broader vision-language alignment perspective of CLIP [41], more recent multimodal document models such as mPLUG-Doc [42] and Vary [43] have further advanced document understanding with stronger multimodal representations.

Our work connects this perspective to cross-dataset adaptation. Rather than proposing a new pre-training objective, we show that the consistency of the supervision signal itself shapes the geometry of the learned representation. By harmonizing heterogeneous annotations before fine-tuning, we obtain detector features that are more compact, more separable, and better aligned with document layout semantics.

## 3 Method

### 3.1 Agentic Harmonization

**Problem formulation.** Consider a collection of $K$ source datasets $\{\mathcal{D}_1, ..., \mathcal{D}_K\}$, where each dataset $\mathcal{D}_k = \{(I_i, \mathcal{A}_i)\}_{i=1}^{N_k}$ consists of $N_k$ document images $I_i$ paired with annotation sets $\mathcal{A}_i = \{(b_j, c_j)\}_{j=1}^{M_i}$, with $b_j \in \mathbb{R}^4$ denoting a bounding box and $c_j$ its category label. Even when the label spaces across datasets share overlapping category names, the spatial semantics encoded in $b_j$ may differ substantially: one dataset may define paragraph-level regions encompassing surrounding whitespace, while another annotates fine-grained text blocks aligned with OCR extraction boundaries. Similar discrepancies arise for tables, captions, forms, and list structures. Naïvely merging $\{\mathcal{D}_k\}$ into a single training set introduces conflicting spatial supervision, as the detector receives inconsistent definitions of object boundaries for semantically equivalent categories.

**Overview.** We propose an agentic harmonization stage that transforms each annotation set $\mathcal{A}_i$ into a harmonized counterpart $\hat{\mathcal{A}}_i$ prior to detector fine-tuning:

$$\hat{\mathcal{A}}_i = \mathcal{F}(I_i, \mathcal{A}_i; \mathcal{R}), \tag{1}$$

where $\mathcal{F}$ is a VLM agent operating on each page independently and $\mathcal{R}$ is a structured rule set that encodes the *target* annotation standard: the spatial and semantic conventions of the detector's native label definitions. We derive $\mathcal{R}$ by analyzing the annotation philosophy of the target dataset to capture how each category is spatially defined. Given a document image $I_i$ and its *source* annotations $\mathcal{A}_i$ from one of the $K$ input datasets, the agent reasons over the visual layout and the source annotations to determine the adjustments needed, for example refining bounding boxes or reassigning categories, so that $\hat{\mathcal{A}}_i$ are transformed to the conventions prescribed by $\mathcal{R}$.

**Constrained transformation.** For a given page $i$, let the source annotations be

$$\mathcal{A}_i = \{a_1, ..., a_{M_i}\}, \quad a_j = (b_j, c_j). \tag{2}$$

The agent decides, for each source annotation, whether it should be kept on its own or merged with one or more neighboring annotations. This amounts to grouping the $M_i$ source annotations into $G_i$ disjoint groups,

$$\Pi_i = \{S_1, ..., S_{G_i}\}, \quad \bigcup_{r=1}^{G_i} S_r = \{1, ..., M_i\}, \quad S_r \cap S_s = \emptyset \text{ for } r \neq s, \tag{3}$$

and then transforming each group into a single harmonized annotation according to the target conventions $\mathcal{R}$:

$$\hat{a}_r = M_{\mathcal{R}}(I_i, \{a_j \mid j \in S_r\}), \quad \hat{\mathcal{A}}_i = \{\hat{a}_1, ..., \hat{a}_{G_i}\}. \tag{4}$$

A group containing a single annotation undergoes individual adjustment (e.g., boundary refinement or category reassignment), while a group containing multiple annotations is first merged and then adjusted. In both cases the same operator $M_\mathcal{R}$ applies, making the transformation uniform regardless of group size.

**Design constraints.** The harmonization agent is governed by two core constraints that ensure consistency and traceability:

1. **Visual-grounded reasoning.** The agent conditions on the document image $I_i$ before modifying any annotation, ensuring that spatial decisions are grounded in observed page structure rather than derived from label names alone.

2. **Conservation of annotations.** The mapping $\mathcal{A}_i \to \hat{\mathcal{A}}_i$ is constrained such that every input annotation is accounted for exactly once in the harmonized output, either preserved or merged with spatially and semantically compatible neighbors, or correct misaligned boundaries to snap to visual edges. While the architecture supports dividing under-segmented regions (e.g., separating a caption from a figure), this rule is excluded from the current study to maintain a controlled experimental scope. The agent may not hallucinate new regions or discard existing ones.

**Harmonized training.** The harmonized datasets $\hat{\mathcal{D}}_k = \{(I_i, \hat{\mathcal{A}}_i)\}$ are merged into a unified training set with consistent spatial supervision. This removes the conflicting boundary definitions introduced by heterogeneous annotation conventions and yields a more stable optimization landscape for mixed-dataset fine-tuning.

## 3.2 Training and Evaluation

### 3.2.1 Agentic Harmonization

We evaluate whether harmonization improves adaptation relative to conventional fine-tuning on heterogeneous datasets. Our base detector is an **IBM Docling Heron** [20] model, specifically an RT-DETR v2 layout model with a ResNet-50vd backbone, 300 object queries, a 6-layer decoder, and contrastive denoising with 100 denoising queries.

We fine-tune for 50 epochs using AdamW with a base learning rate of $3 \times 10^{-4}$, a backbone learning rate of $3 \times 10^{-5}$, 2,000 warmup steps, cosine annealing, and gradient clipping at 0.1. Experiments are conducted on 2 NVIDIA A100 80GB GPUs.

Because the source datasets define incompatible category taxonomies, standard per-class detection metrics such as mAP cannot be computed fairly across training regimes. We therefore evaluate each model through end-to-end file transformation quality using the SCORE framework [44], which measures how accurately the full document conversion pipeline reproduces ground-truth document structure. This approach captures the practical impact of detection improvements on the downstream task that users ultimately care about: faithful document conversion.

SCORE captures content fidelity, table structure accuracy, spatial consistency, and reading order correctness in a single evaluation suite, providing a holistic assessment that goes beyond pure box overlap. All models are evaluated on SCORE-Bench[1], a standardized benchmark of diverse document types designed for reproducible end-to-end evaluation.

---

[1]https://huggingface.co/datasets/unstructuredio/SCORE-Bench

### 3.2.2 Representation Analysis

We hypothesize that annotation inconsistency distorts the learned feature space, we go beyond standard evaluation and analyze the detector's latent geometry.

**Feature extraction** Let $\boldsymbol{h}_{\text{dec}}(I)$ denote the feature representations extracted from the detector after the decoder stage. These post-decoder features reflect the refined object queries most directly tied to final detection behavior and category assignment. By analyzing this space, we can observe how harmonization affects the final semantic clustering of layout elements compared to naïve merging of inconsistent datasets.

**Evaluation metrics** We assess representation quality using three complementary tools:

- **UMAP Projections:** To visualize global class organization and cross-class overlap in the post-decoder embedding space.
- **Silhouette Score:** To evaluate the compactness of each class relative to its separation from neighboring classes; higher scores indicate more distinct semantic boundaries.
- **Neighborhood Purity:** Calculated as the fraction of same-class samples among each point's $k$ nearest neighbors to measure local semantic consistency.

For representation analysis, we extract detector features from the pretrained Heron model, the naïvely fine-tuned baseline, and our harmonized model under matched evaluation settings. All visualization and clustering metrics are computed on a fixed set of evaluation pages to isolate the effect of the training strategy on representation geometry. We focus our analysis on the post-decoder representation space, as these features are most directly linked to the final detection predictions and reflect the model's ultimate semantic organization.

## 4 Datasets

Our work builds on the Heron family of document layout models [20], which fine-tune RT-DETRv2 on large, heterogeneous document corpora. Although the full set of training datasets used by Heron is not publicly available, one of its core components, DocLayNet [8], is openly released. We sourced the 25,000-image subset from [8] and used it alongside the Unstructured proprietary dataset to study the annotation discrepancies that arise when combining independently curated document sources. Table 1 summarizes the scale and composition of each split.

| Split | Images | Annotations | Avg Ann/Img | Avg Size (W×H) |
| --- | --- | --- | --- | --- |
| Unstructured Train | 47,744 | 810,644 | 17.0 | 1765×2179 |
| Unstructured Val | 13,642 | 230,945 | 16.9 | 1766×2203 |
| DocLayNet 25k | 25,000 | 328,756 | 13.2 | 1025×1025 |

Table 1: Overview of dataset splits used in our experiments.

**Docling base model taxonomy.** Our base detector is initialized from the Docling Heron [20], which defines a canonical taxonomy of 17 document element categories. Table 2 shows the distribution of annotations across train, validation, and test splits, totaling 2,805,191 annotations. The taxonomy is dominated by `Text`and `List-item`, `Section-header` and `Picture`. Several categories are relatively rare: `Document Index`, `Checkbox-Selected`, and `Code`.

| Category | Train | Val | Test | Total |
|---|---:|---:|---:|---:|
| Caption | 37,680 | 4,252 | 3,860 | 45,792 |
| Checkbox-Selected | 3,071 | 455 | 451 | 3,977 |
| Checkbox-Unselected | 45,260 | 5,827 | 6,261 | 57,348 |
| Code | 6,185 | 760 | 727 | 7,672 |
| Document Index | 1,587 | 179 | 208 | 1,974 |
| Footnote | 9,818 | 1,168 | 1,200 | 12,186 |
| Form | 12,521 | 1,625 | 1,566 | 15,712 |
| Formula | 29,101 | 2,704 | 2,923 | 34,728 |
| Key-Value Region | 20,649 | 2,665 | 2,683 | 25,997 |
| List-item | 421,845 | 47,552 | 47,426 | 516,823 |
| Page-footer | 107,761 | 11,321 | 11,304 | 130,386 |
| Page-header | 92,304 | 10,636 | 10,251 | 113,191 |
| Picture | 128,603 | 14,697 | 14,530 | 157,830 |
| Section-header | 293,021 | 34,368 | 31,212 | 358,601 |
| Table | 31,877 | 2,964 | 2,941 | 37,782 |
| Text | 1,042,044 | 118,863 | 115,479 | 1,276,386 |
| Title | 7,120 | 838 | 848 | 8,806 |
| **Total** | **2,290,447** | **260,874** | **253,870** | **2,805,191** |

Table 2: Distribution of the 17 canonical document element categories in the Docling base model training data.

**Taxonomy mismatch.** Neither the Unstructured nor DocLayNet label space aligns directly with the Docling canonical taxonomy. In our sample, the `Footnote` category is absent, resulting in a 10-category taxonomy for DocLayNet. Table 3 shows the class-by-class correspondence between the Unstructured and DocLayNet datasets. Of the 16 Unstructured categories, only 8 have direct counterparts in DocLayNet 25k: `paragraph`, `subheading`, `title`, `table`, `image`, `formulas`, `page_header`, and `page_footer`. The Unstructured set contains 8 additional categories absent from DocLayNet, including `checkbox`, `form`, `code_snippet`, `page_number`, and `other`, while DocLayNet introduces `Caption` (5.3%) and `List-item` (11.4%) that have no Unstructured equivalent. This partial overlap means that naively merging the two sources introduces categories that the other dataset has never annotated, creating both missing-label and extra-label conflicts that must be reconciled against the Docling canonical taxonomy.

| Unstructured | DocLayNet | Unstr. Count | Unstr. % | Doc. Count | Doc. % |
| --- | --- | --- | --- | --- | --- |
| paragraph | Text | 413,074 | 51.0% | 140,107 | 42.6% |
| subheading | Section-header | 86,972 | 10.7% | 36,085 | 11.0% |
| title | Title | 15,213 | 1.9% | 4,131 | 1.3% |
| table | Table | 19,965 | 2.5% | 18,300 | 5.6% |
| image | Picture | 47,345 | 5.8% | 18,885 | 5.7% |
| formulas | Formula | 4,625 | 0.6% | 18,094 | 5.5% |
| page_header | Page-header | 19,185 | 2.4% | 15,637 | 4.8% |
| page_footer | Page-footer | 20,678 | 2.5% | 22,609 | 6.9% |
| checkbox | — | 3,358 | 0.4% | — | — |
| checkbox_checked | — | 1,621 | 0.2% | — | — |
| code_snippet | — | 1,133 | 0.1% | — | — |
| form | — | 1,850 | 0.2% | — | — |
| other | — | 137,201 | 16.9% | — | — |
| page_number | — | 37,354 | 4.6% | — | — |
| radio_button | — | 870 | 0.1% | — | — |
| radio_button_checked | — | 200 | 0.0% | — | — |
| — | Caption | — | — | 17,370 | 5.3% |
| — | List-item | — | — | 37,538 | 11.4% |

Table 3: Taxonomy mapping between the Unstructured and DocLayNet 25k datasets. Only 8 of the 16 Unstructured categories have direct counterparts in DocLayNet. Rows below the midrule show dataset-specific categories with no match in the other source.

**Spatial discrepancy.** Even among the 8 shared categories, annotation conventions differ substantially. Table 4 reports bounding box statistics for matched classes. The Unstructured dataset uses high-resolution page images ($\sim 1765 \times 2179$), while DocLayNet renders pages at a fixed $1025 \times 1025$. After accounting for this resolution gap, the spatial extent of annotations still diverges: `table` regions in the Unstructured set are $5.2\times$ larger by area, `page_footer` boxes are $6.3\times$ larger, and `subheading` boxes are $4.7\times$ larger. In contrast, `formulas` boxes in DocLayNet are wider than their Unstructured counterparts ($0.74\times$ width ratio), and `page_header` boxes are taller in DocLayNet ($0.80\times$ height ratio). These asymmetries reflect fundamentally different decisions about whitespace inclusion, grouping granularity, and annotation scope.

| Class | Unstr. W | Unstr. H | Unstr. Area | Doc Area | W Ratio | Area Ratio |
|---|---|---|---|---|---|---|
| paragraph | 832.7 | 122.5 | 102,006 | 36,186 | 1.64× | 2.82× |
| subheading | 469.4 | 42.8 | 20,090 | 4,276 | 1.84× | 4.70× |
| title | 697.7 | 83.0 | 57,909 | 13,430 | 1.70× | 4.31× |
| table | 1294.6 | 628.9 | 814,174 | 155,576 | 1.91× | 5.23× |
| image | 635.8 | 420.5 | 267,354 | 94,073 | 1.53× | 2.84× |
| formulas | 326.9 | 93.8 | 30,663 | 21,057 | 0.74× | 1.46× |
| page_header | 374.0 | 34.8 | 13,015 | 7,556 | 2.16× | 1.72× |
| page_footer | 413.5 | 30.9 | 12,777 | 2,024 | 2.96× | 6.31× |

Table 4: Average absolute bounding box dimensions (pixels) for matched categories. Unstructured images are $\sim 1765 \times 2179$; DocLayNet images are $1025 \times 1025$. W Ratio = Unstr. width / Doc width; Area Ratio = Unstr. area / Doc area.

**Class distribution imbalance.** Beyond spatial differences, the two datasets weight shared categories very differently. `paragraph` accounts for 42.6% of DocLayNet annotations and 51.0% of the Unstructured set: an 8.3 percentage point gap. Both datasets are dominated by paragraph annotations, but the Unstructured set is even more skewed toward this single category. Similarly, `formulas` represents 5.5% of DocLayNet but only 0.6% of the Unstructured set, and `table` is 5.6% versus 2.5%. When these datasets are merged without harmonization, the detector receives supervision in which the same category label carries different implicit frequency priors, further compounding the spatial inconsistencies described above.

**Target taxonomy.** To reconcile these mismatches, we define a unified target taxonomy of 17 categories that covers layout elements observed across both sources: `paragraph`, `subheading`, `title`, `table`, `image`, `figure_caption`, `list_item`, `formulas`, `page_header`, `page_footer`, `page_number`, `checkbox`, `checkbox_checked`, `form`, `form_key_values`, `code_snippet`, and `other`. This taxonomy retains the fine-grained form and checkbox categories from the Unstructured set while incorporating `figure_caption` and `list_item` from DocLayNet. All subsequent harmonization decisions map source annotations into this common label space.

**Implications.** Taken together, these three dimensions of discrepancy, taxonomy, spatial extent, and class distribution point to the challenge that motivates our agentic harmonization approach (Section 3). The problem is not simply that the datasets come from different domains; it is that even their shared categories encode conflicting definitions of what constitutes a correct annotation. Training on such data without reconciliation forces the detector to absorb these conflicts as noise, degrading both detection quality and the coherence of learned layout representations. A detailed per-class breakdown is provided in Section 7.

# 5 Experimental Results

We evaluate whether agentic harmonization improves cross-dataset adaptation under heterogeneous annotation semantics. We compare three training regimes throughout the study:
- a pretrained detector (i.e., Docling Heron) without additional adaptation,
- a detector obtained by naïve fine-tuning on mixed datasets, and
- a detector fine-tuned on the harmonized training set.

As mentioned in Section 3.2.1, standard mAP metrics cannot be computed fairly across the three training regimes; we therefore evaluate each model through end-to-end file transformation quality, measuring how accurately the full document conversion pipeline reproduces the ground-truth document structure. Our analysis is organized along two axes. First, we measure end-to-end file transformation quality by plugging each detector into the Unstructured **High Fidelity Transform** pipeline. Second, we analyze the geometry of the learned representation space using low-dimensional visualizations and clustering-based metrics. This combination allows us to connect supervision consistency during training to both downstream utility and representation quality.

## 5.1 End-to-End File Transformation Results

Table 5 reports results across 17 metrics spanning content fidelity, table structure, and spatial consistency (see Section 7.1 for definitions). The central question is whether simply exposing the detector to more data is sufficient, or whether heterogeneous annotation semantics undermine the benefit of mixed-dataset fine-tuning.

Naïve fine-tuning on the mixed dataset degrades pipeline quality on most metrics relative to the pretrained Docling baseline. The page-level TEDS falls from 0.778 to 0.733, and cell-level content accuracy decreases from 0.767 to 0.715. This is consistent with the expected effect of conflicting supervision: when semantically similar labels correspond to different spatial definitions across datasets, the detector absorbs these conflicts as noise, weakening the consistency of the learned spatial prior and propagating errors downstream.

In contrast, fine-tuning with agentic harmonization produces consistent improvements over both the pretrained baseline and naïve mixed-dataset fine-tuning. The harmonized model achieves the best score on 14 of 17 metrics. Detection F-score improves from 0.860 to 0.883, with corresponding gains in both precision (0.868 → 0.885) and recall (0.858 → 0.887). Table structure metrics show particularly strong gains: table TEDS rises from 0.800 to 0.814 and cell-level index accuracy from 0.764 to 0.778. Bounding box overlap metrics also improve, with the mean IoU of overlapping pairs dropping from 0.043 to 0.016, indicating cleaner spatial predictions. These results support the main premise of the paper: increasing training volume alone is insufficient when the underlying supervision is inconsistent, whereas resolving annotation mismatches before training yields more effective adaptation.

| Metric | Heron | Naïve | Harmonized |
|---|---|---|---|
| adjusted_NED ↑ | 0.870 | 0.871 | **0.872** |
| NED ↑ | 0.866 | 0.867 | **0.868** |
| detection_f ↑ | 0.860 | 0.858 | **0.883** |
| detection_precision ↑ | 0.868 | 0.868 | **0.885** |
| detection_recall ↑ | 0.858 | 0.858 | **0.887** |
| page_teds_corrected ↑ | 0.778 | 0.733 | **0.790** |
| table_teds ↑ | 0.800 | 0.750 | **0.814** |
| table_teds_corrected ↑ | 0.777 | 0.730 | **0.792** |
| cell_level_content_acc ↑ | 0.767 | 0.715 | **0.771** |
| cell_level_index_acc ↑ | 0.764 | 0.725 | **0.778** |
| shifted_cell_content_acc ↑ | 0.806 | 0.751 | **0.807** |
| element_alignment ↑ | **0.585** | 0.549 | 0.579 |
| percent_tokens_found ↑ | **0.947** | 0.941 | 0.946 |
| percent_tokens_added ↓ | 0.052 | **0.046** | 0.049 |
| bbox_max_iou ↓ | 0.079 | 0.045 | **0.035** |
| bbox_mean_iou ↓ | 0.043 | 0.019 | **0.016** |
| bbox_num_overlapping_pairs ↓ | 4.058 | 4.460 | **4.036** |

Table 5: End-to-end file transformation results. See Section 7.1 for more details about the metrics presented.

## 5.2 Representation Analysis

To better understand why harmonized training improves end-to-end detection quality, we examine the structure of the detector's latent space. Our hypothesis is that inconsistent supervision does not merely affect box predictions; it also distorts the internal geometry of the learned representation. We therefore compare the pretrained, naïvely fine-tuned, and harmonized detectors using embedding visualizations and clustering metrics.

### 5.2.1 Embedding Space Structure

Figure 3 visualizes the post-decoder embedding space under each training regime. The pretrained model already forms partially organized clusters, but several classes remain entangled and the global structure is only moderately separable. After naïve fine-tuning, the geometry becomes less coherent: clusters fragment, class boundaries blur, and inter-class mixing increases. This is consistent with the expected effect of conflicting supervision, where the model is encouraged to map visually similar but inconsistently annotated elements to overlapping regions of feature space.

By contrast, harmonized training yields a visibly more structured post-decoder space with clearer organization, more compact clusters, and stronger margins between semantically distinct layout categories.

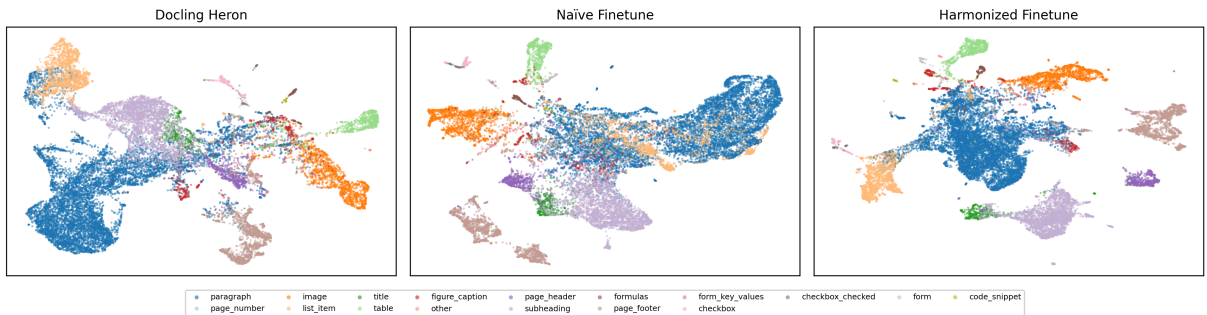

Figure 3: Post-decoder embedding space under the three training regimes. Compared with the pretrained and naïvely fine-tuned models, the harmonized model exhibits tighter clusters and clearer inter-class boundaries, indicating a more transferable layout representation.

To examine whether these trends hold for the most challenging categories, Figure 4 focuses on several commonly confused class pairs. The naïvely fine-tuned model often increases overlap for ambiguous pairs such as `paragraph` versus `list_item` and `title` versus `subheading`. In contrast, the harmonized model produces noticeably clearer separation for these pairs, suggesting that resolving annotation inconsistencies sharpens the corresponding decision boundaries in the learned feature space.

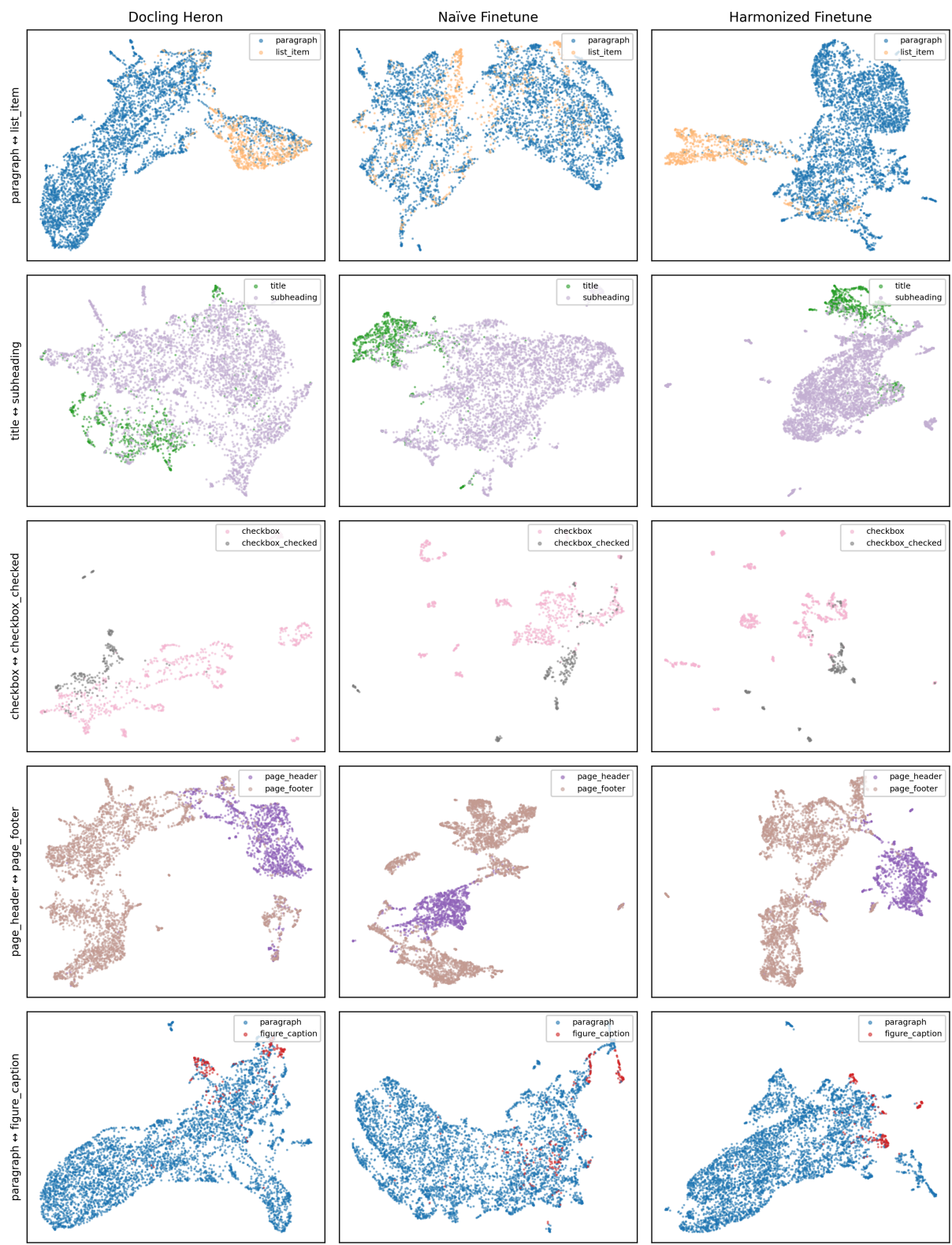

Figure 4: Pairwise UMAP projections for commonly confused layout categories. Harmonized training improves separation for several ambiguous class pairs, while naïve fine-tuning often increases overlap and boundary instability.

### 5.2.2 Global Cluster Separation

We further quantify the global structure of the representation space using per-class silhouette scores computed on the detector embeddings. The silhouette score measures how compact each class cluster is relative to its separation from neighboring classes and therefore provides a useful summary of both intra-class consistency and inter-class discrimination.

Because Heron's 17 canonical categories (Table 2) do not exactly match our 17-class target taxonomy one-to-one, we remap Heron's predictions into the target label space by semantic correspondence before computing silhouette scores, so that all three models are analyzed in the same label space. The mapping is straightforward for most categories (e.g., Heron `Text` → `paragraph`, `Section-header` → `subheading`, `Picture` → `image`, `Caption` → `figure_caption`, `List-item` → `list_item`). Heron's `Footnote` and `Document Index` categories do not have direct counterparts in our target taxonomy; we assign both to the `other` class. This remapping is used only for the representation analysis; the end-to-end evaluation in Section 5.1 is unaffected, as it is computed via the downstream conversion pipeline.

Figure 5 shows that naïve fine-tuning does not consistently improve this metric and, in several cases, degrades it, particularly for classes that are already difficult to disambiguate under mixed annotation standards. In contrast, harmonized training improves post-decoder silhouette scores across many categories and reduces the number of classes with negative or near-zero separation. The largest gains appear in categories that are especially sensitive to annotation ambiguity, which aligns with the qualitative trends observed in the embedding plots.

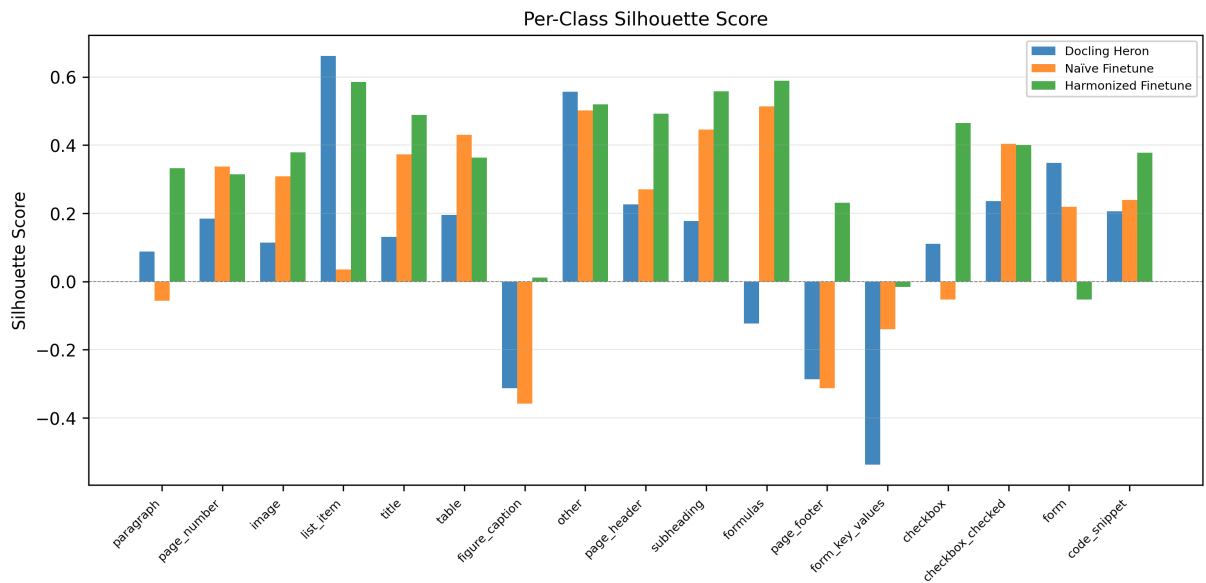

Figure 5: Per-class silhouette scores for post-decoder representations. Harmonized training improves global cluster separation across many classes compared to both the pretrained and naïvely fine-tuned models.

We also compute neighborhood purity using the fraction of the same class samples among each point's $k = 100$ nearest neighbors. As shown in Figure 6, all three models retain high local purity, but the harmonized model reaches the strongest and most consistent post-decoder neighborhood structure. This result complements the silhouette analysis: local neighborhoods remain semantically coherent across settings, but harmonized training improves the larger-scale arrangement of the feature space and yields better global separability.

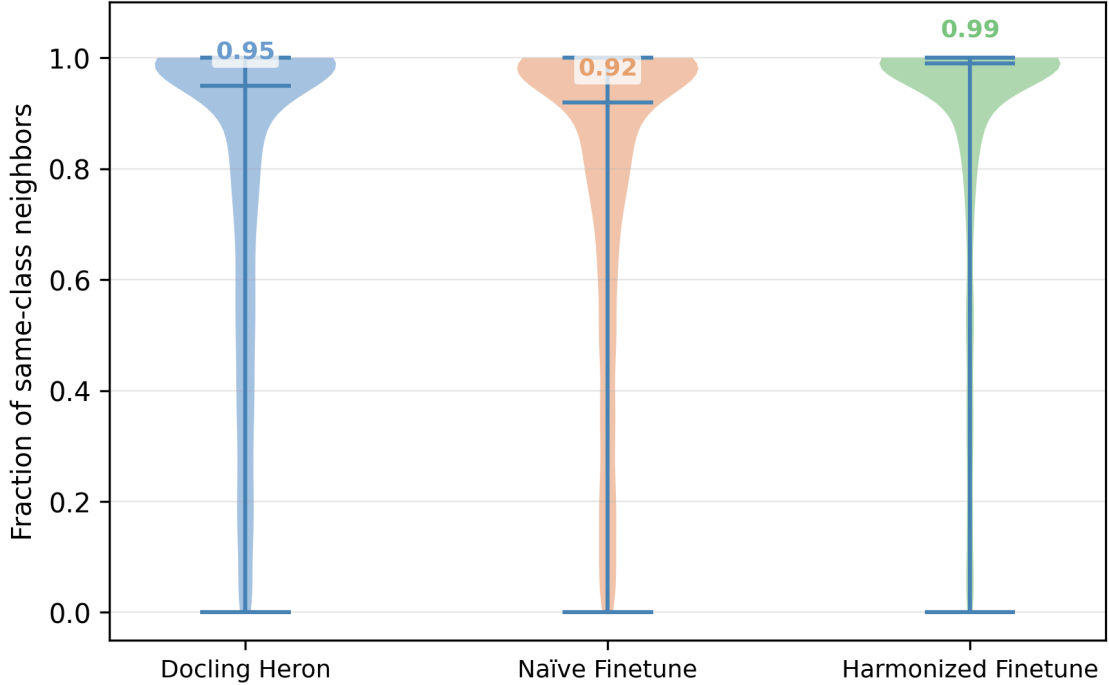

Figure 6: Neighborhood purity for post-decoder embeddings using $k = 100$ nearest neighbors. All models maintain strong local consistency, while the harmonized model shows the strongest neighborhood structure.

Taken together, these analyses indicate that harmonization improves the internal organization of the detector's post-decoder representation space, not only its output-level predictions. This provides a coherent explanation for the end-to-end improvements observed in Section 5.1: by aligning label semantics and bounding box granularity before training, the detector learns a feature space that is more compact, more separable, and better suited for the downstream document conversion.

## 6 Discussions & Future Work

The results of this study indicate that annotation mismatch is a consequential source of error in multi-source detector adaptation. When datasets employ different spatial conventions for semantically related layout elements, naïve joint fine-tuning exposes the model to inconsistent supervision. In the setting considered here, harmonization mitigates this effect by aligning source annotations to a common target standard prior to training. The observed improvements in detection performance and representation structure suggest that annotation consistency influences not only output quality but also the internal organization of the learned feature space.

Several ongoing research directions should be considered in future work. First, the current harmonization procedure operates under a no-splitting regime: source annotations may be preserved, merged, or boundary-adjusted, but they are not divided into multiple target instances. While this formulation covers the majority of practical annotation mismatches, certain cases involving under-segmented composite regions would benefit from more expressive transformations such as one-to-many splitting. Second, the empirical study is confined to document layout detection with partially overlapping taxonomies and a fixed target annotation standard. Although this setting is practically relevant, the extent to which the present findings generalize to other domains, annotation ontologies, or visual regimes remains to be established. Third, harmonization is performed as a preprocessing stage, independently of detector optimization. As a consequence, potential interactions between annotation transformation and representation learning are not explored in the current formulation.

These limitations define several natural directions for future work. An immediate extension would be to permit one-to-many transformations, thereby accommodating annotation schemes that differ through both over-segmentation and under-segmentation. It would also be useful to investigate uncertainty-aware harmonization procedures that identify low-confidence transformations for selective review rather than automatically resolving them in all cases. A further direction is joint or iterative optimization, in which harmonization and detector adaptation are coupled rather than treated as sequential stages. Such formulations may clarify whether supervision alignment and representation learning can beneficially inform one another during training.

More broadly, the present results suggest that supervision consistency is an important variable in dataset integration for structured vision tasks. In settings where detectors serve as upstream components for downstream parsing, retrieval, or multimodal reasoning, annotation harmonization may provide a principled mechanism for reducing inter-dataset incompatibilities before model adaptation. Evaluating this hypothesis across broader collections of datasets and task settings remains an important direction for future work.

# Bibliography

[1] J. Lambert, Z. Liu, O. Sener, J. Hays, and V. Koltun, "MSeg: A Composite Dataset for Multi-domain Semantic Segmentation," in *IEEE/CVF Conference on Computer Vision and Pattern Recognition (CVPR)*, 2020. [Online]. Available: https://arxiv.org/abs/2112.13762

[2] Z. Chen *et al.*, "Dynamic Supervisor for Cross-dataset Object Detection," *Neurocomputing*, 2022, [Online]. Available: https://arxiv.org/abs/2204.00183

[3] Y.-H. Liao, D. Acuna, R. Mahmood, J. Lucas, V. Prabhu, and S. Fidler, "Transferring Labels to Solve Annotation Mismatches Across Object Detection Datasets," in *International Conference on Learning Representations (ICLR)*, 2024. [Online]. Available: https://openreview.net/forum?id=ChHx5ORqF0

[4] J. Kirkpatrick *et al.*, "Overcoming Catastrophic Forgetting in Neural Networks," *Proceedings of the National Academy of Sciences (PNAS)*, vol. 114, no. 13, pp. 3521–3526, 2017, [Online]. Available: https://arxiv.org/abs/1612.00796

[5] T. Feng, M. Wang, and H. Yuan, "Overcoming Catastrophic Forgetting in Incremental Object Detection via Elastic Response Distillation," in *IEEE/CVF Conference on Computer Vision and Pattern Recognition (CVPR)*, 2022. [Online]. Available: https://arxiv.org/abs/2204.02136

[6] X. Zhong, J. Tang, and A. J. Yepes, "PubLayNet: Largest Dataset Ever for Document Layout Analysis," in *International Conference on Document Analysis and Recognition (ICDAR)*, 2019. [Online]. Available: https://arxiv.org/abs/1908.07836

[7] M. Li *et al.*, "DocBank: A Benchmark Dataset for Document Layout Analysis," in *International Conference on Computational Linguistics (COLING)*, 2020. [Online]. Available: https://arxiv.org/abs/2006.01038

[8] B. Pfitzmann, C. Auer, M. Dolfi, A. S. Nassar, and P. W. J. Staar, "DocLayNet: A Large Human-Annotated Dataset for Document-Layout Segmentation," in *Proceedings of the 28th ACM SIGKDD Conference on Knowledge Discovery and Data Mining*, 2022, pp. 3743–3751. doi: 10.1145/3534678.3539043.

[9] S. Ren, K. He, R. Girshick, and J. Sun, "Faster R-CNN: Towards Real-Time Object Detection with Region Proposal Networks," in *Advances in Neural Information Processing Systems (NeurIPS)*, 2015. [Online]. Available: https://arxiv.org/abs/1506.01497

[10] J. Redmon, S. Divvala, R. Girshick, and A. Farhadi, "You Only Look Once: Unified, Real-Time Object Detection," in *IEEE/CVF Conference on Computer Vision and Pattern Recognition (CVPR)*, 2016. [Online]. Available: https://arxiv.org/abs/1506.02640

[11] Z. Ge, S. Liu, F. Wang, Z. Li, and J. Sun, "YOLOX: Exceeding YOLO Series in 2021," *arXiv preprint arXiv:2107.08430*, 2021, [Online]. Available: https://arxiv.org/abs/2107.08430

[12] G. Jocher, A. Chaurasia, and J. Qiu, "YOLOv8." [Online]. Available: https://github.com/ultralytics/ultralytics

[13] N. Carion, F. Massa, G. Synnaeve, N. Usunier, A. Kirillov, and S. Zagoruyko, "End-to-End Object Detection with Transformers," in *European Conference on Computer Vision (ECCV)*, 2020. [Online]. Available: https://arxiv.org/abs/2005.12872

[14] X. Zhu, W. Su, L. Lu, B. Li, X. Wang, and J. Dai, "Deformable DETR: Deformable Transformers for End-to-End Object Detection," in *International Conference on Learning Representations (ICLR)*, 2021. [Online]. Available: https://arxiv.org/abs/2010.04159

[15] H. Zhang *et al.*, "DINO: DETR with Improved DeNoising Anchor Boxes for End-to-End Object Detection," in *International Conference on Learning Representations (ICLR)*, 2023. [Online]. Available: https://arxiv.org/abs/2203.03605

[16] Y. Zhao *et al.*, "DETRs Beat YOLOs on Real-time Object Detection." 2023.

[17] W. Lv, Y. Zhao, Q. Chang, K. Huang, G. Wang, and Y. Liu, "RT-DETRv2: Improved Baseline with Bag-of-Freebies for Real-Time Detection Transformer," *arXiv preprint arXiv:2407.17140*, 2024, [Online]. Available: https://arxiv.org/abs/2407.17140

[18] C. Da, C. Luo, Q. Zheng, and C. Yao, "VGT: Vision Grid Transformer for Document Layout Analysis," in *IEEE/CVF International Conference on Computer Vision (ICCV)*, 2023. [Online]. Available: https://arxiv.org/abs/2308.14978

[19] Z. Zhao, H. Kang, B. Wang, and C. He, "DocLayout-YOLO: Enhancing Document Layout Analysis through Diverse Synthetic Data and Global-to-Local Adaptive Perception," *arXiv preprint arXiv:2410.12628*, 2024, [Online]. Available: https://arxiv.org/abs/2410.12628

[20] N. Livathinos, C. Auer, A. Nassar, and others, "Advanced Layout Analysis Models for Docling," *arXiv preprint arXiv:2509.11720*, 2025, [Online]. Available: https://arxiv.org/abs/2509.11720

[21] S. J. Pan and Q. Yang, "A Survey on Transfer Learning," *IEEE Transactions on Knowledge and Data Engineering*, vol. 22, no. 10, pp. 1345–1359, 2010, [Online]. Available: https://ieeexplore.ieee.org/document/5288526

[22] W. Li, F. Li, Y. Luo, P. Wang, and J. Sun, "Deep Domain Adaptive Object Detection: A Survey," in *IEEE Symposium Series on Computational Intelligence (SSCI)*, 2020. [Online]. Available: https://arxiv.org/abs/2002.06797

[23] Y. Chen, W. Li, C. Sakaridis, D. Dai, and L. Van Gool, "Domain Adaptive Faster R-CNN for Object Detection in the Wild," in *IEEE/CVF Conference on Computer Vision and Pattern Recognition (CVPR)*, 2018. [Online]. Available: https://arxiv.org/abs/1803.03243

[24] K. Saito, Y. Ushiku, T. Harada, and K. Saenko, "Strong-Weak Distribution Alignment for Adaptive Object Detection," in *IEEE/CVF Conference on Computer Vision and Pattern Recognition (CVPR)*, 2019. [Online]. Available: https://arxiv.org/abs/1812.04798

[25] C.-C. Hsu, Y.-H. Tsai, Y.-Y. Lin, and M.-H. Yang, "Every Pixel Matters: Center-Aware Feature Alignment for Domain Adaptive Object Detector," in *European Conference on Computer Vision (ECCV)*, 2020. [Online]. Available: https://arxiv.org/abs/2008.08574

[26] G. Zhao, G. Li, R. Xu, and L. Lin, "Collaborative Training between Region Proposal Localization and Classification for Domain Adaptive Object Detection," in *European Conference on Computer Vision (ECCV)*, 2020. [Online]. Available: https://arxiv.org/abs/2009.08119

[27] X. Wang, Z. Cai, D. Gao, and N. Vasconcelos, "Towards Universal Object Detection by Domain Attention," in *IEEE/CVF Conference on Computer Vision and Pattern Recognition (CVPR)*, 2019. [Online]. Available: https://arxiv.org/abs/1904.04402

[28] X. Zhou, R. Girdhar, A. Joulin, P. Krähenbühl, and I. Misra, "Detecting Twenty-Thousand Classes Using Image-Level Supervision," in *European Conference on Computer Vision (ECCV)*, 2022. [Online]. Available: https://arxiv.org/abs/2201.02605

[29] M. Kennerley, A. I. Aviles-Rivero, C.-B. Schönlieb, and R. T. Tan, "Bridging Annotation Gaps: Transferring Labels to Align Object Detection Datasets," *arXiv preprint arXiv:2506.04737*, 2025, [Online]. Available: https://arxiv.org/abs/2506.04737

[30] L. H. Li *et al.*, "Grounded Language-Image Pre-Training," in *IEEE/CVF Conference on Computer Vision and Pattern Recognition (CVPR)*, 2022. [Online]. Available: https://arxiv.org/abs/2112.03857

[31] S. Liu *et al.*, "Grounding DINO: Marrying DINO with Grounded Pre-Training for Open-Set Object Detection," *arXiv preprint arXiv:2303.05499*, 2023, [Online]. Available: https://arxiv.org/abs/2303.05499

[32] J. Hoffman *et al.*, "LSDA: Large Scale Detection Through Adaptation," in *Advances in Neural Information Processing Systems (NeurIPS)*, 2014. [Online]. Available: https://arxiv.org/abs/1407.5035

[33] H. Rasheed, M. Maaz, M. U. Khattak, S. H. Khan, and F. S. Khan, "Bridging the Gap between Object and Image-Level Representations for Open-Vocabulary Detection," in *Advances in Neural Information Processing Systems (NeurIPS)*, 2022. [Online]. Available: https://arxiv.org/abs/2207.03482

[34] S. Tewes, Y. Chen, O. Moured, J. Zhang, and R. Stiefelhagen, "SFDLA: Source-Free Document Layout Analysis," in *International Conference on Document Analysis and Recognition (ICDAR)*, 2025. [Online]. Available: https://arxiv.org/abs/2503.18742

[35] J. Wang, K. Hu, Z. Zhong, L. Sun, and Q. Huo, "Detect-Order-Construct: A Tree Construction based Approach for Hierarchical Document Structure Analysis," *Pattern Recognition*, 2024, [Online]. Available: https://www.sciencedirect.com/science/article/pii/S0031320324005879

[36] Y. Xu, M. Li, L. Cui, S. Huang, F. Wei, and M. Zhou, "LayoutLM: Pre-training of Text and Layout for Document Image Understanding," in *ACM SIGKDD International Conference on Knowledge Discovery and Data Mining (KDD)*, 2020. [Online]. Available: https://arxiv.org/abs/1912.13318

[37] S. Appalaraju, B. Jasani, B. U. Kota, Y. Xie, and R. Manmatha, "DocFormer: End-to-End Transformer for Document Understanding," in *IEEE/CVF International Conference on Computer Vision (ICCV)*, 2021. [Online]. Available: https://arxiv.org/abs/2106.11539

[38] J. Li, Y. Xu, T. Lv, L. Cui, C. Zhang, and F. Wei, "DiT: Self-Supervised Pre-Training for Document Image Transformer," in *ACM International Conference on Multimedia (ACM MM)*, 2022. [Online]. Available: https://arxiv.org/abs/2203.02378

[39] G. Kim *et al.*, "OCR-Free Document Understanding Transformer," in *European Conference on Computer Vision (ECCV)*, 2022. [Online]. Available: https://arxiv.org/abs/2111.15664

[40] Y. Huang, T. Lv, L. Cui, Y. Lu, and F. Wei, "LayoutLMv3: Pre-Training for Document AI with Unified Text and Image Masking," in *ACM International Conference on Multimedia (ACM MM)*, 2022. [Online]. Available: https://arxiv.org/abs/2204.08387

[41] A. Radford *et al.*, "Learning Transferable Visual Models from Natural Language Supervision," in *International Conference on Machine Learning (ICML)*, 2021. [Online]. Available: https://arxiv.org/abs/2103.00020

[42] J. Ye *et al.*, "mPLUG-DocOwl: Modularized Multimodal Large Language Model for Document Understanding," *arXiv preprint arXiv:2307.02499*, 2023, [Online]. Available: https://arxiv.org/abs/2307.02499

[43] H. Wei *et al.*, "Vary: Scaling Up the Vision Vocabulary for Large Vision-Language Models," *arXiv preprint arXiv:2312.06109*, 2023, [Online]. Available: https://arxiv.org/abs/2312.06109

[44] R. Li, A. Jimeno Yepes, Y. You, K. Pluciński, M. Operlejn, and C. Wolfe, "SCORE: A Semantic Evaluation Framework for Generative Document Parsing," *arXiv preprint arXiv:2509.19345*, 2025, [Online]. Available: https://arxiv.org/abs/2509.19345

# 7 Appendix

## 7.1 End-to-end file transformation metrics

The metrics reported in Section 5.1 evaluate a model's ability to produce accurate end-to-end document transformations from page images. The implementation is publicly available at[2].

For readability, the metrics can be grouped into three families.

**Detection quality.** The metrics `detection_precision`, `detection_recall`, and `detection_f` measure instance-level detection quality under the matching protocol used by the evaluation pipeline. These metrics summarize whether the detector recovers the correct layout regions and category assignments before downstream document reconstruction.

**Content and structure fidelity.** The metrics `NED` and `adjusted_NED` summarize document-level content fidelity. The TEDS-based metrics `page_teds_corrected`, `table_teds`, and `table_teds_corrected` measure structural similarity between predicted and reference outputs at the page or table level. The cell-level metrics `cell_level_content_acc`, `cell_level_index_acc`, and `shifted_cell_content_acc` quantify the correctness of reconstructed table cells in terms of content recovery and placement. The metric `element_alignment` measures how well extracted elements remain aligned with the reference document structure. Finally, `percent_tokens_found` and `percent_tokens_added` capture token-level under-recovery and over-generation, respectively.

**Spatial consistency.** The additional metrics `bbox_max_iou`, `bbox_mean_iou`, and `bbox_num_overlapping_pairs` are designed to measure residual overlap among predicted regions on the same page. More specifically, `bbox_max_iou` captures the worst-case overlap among overlapping predicted region pairs, `bbox_mean_iou` summarizes the average degree of overlap across such pairs, and `bbox_num_overlapping_pairs` measures how frequently overlapping predictions occur. Lower values indicate cleaner spatial decomposition, with fewer duplicated or excessively overlapping predictions. In the context of this study, these metrics are particularly useful because annotation mismatch often manifests not only as missed detections but also as unstable or redundant spatial extents.

Taken together, these metrics provide a broader view of model quality than conventional detection metrics alone. They allow us to distinguish between models that merely recover more boxes and models that produce cleaner, more structurally faithful document transformations.

---

[2]https://github.com/Unstructured-IO/unstructured-eval-metrics